\documentclass[10pt,twocolumn,letterpaper]{article}

\usepackage{cvpr}
\usepackage{times}
\usepackage{epsfig}
\usepackage{graphicx}
\usepackage{amsmath}
\usepackage{amssymb}
\usepackage{booktabs}
\usepackage{subfig}
\usepackage[export]{adjustbox}
\usepackage{color,soul}

\usepackage{comment}
\usepackage{xfrac}
\usepackage{caption}

\usepackage[pagebackref=true,breaklinks=true,letterpaper=true,colorlinks,bookmarks=false]{hyperref}

\cvprfinalcopy 

\def\cvprPaperID{5346} 

\ifcvprfinal\fi
\begin{document}

\title{DeepConsensus: using the consensus of features from multiple layers to attain robust image classification}

\author{Yuchen Li, Safwan Hossain\thanks{These authors made equal contribution.}, Kiarash Jamali\footnotemark[1], Frank Rudzicz\\
Vector Institute and University of Toronto\\
Suite 710, 661 University Avenue\\
{\tt\small ychnlgy.li@utoronto.ca, \{safwan.hossain, kiarash.jamali\}@mail.utoronto.ca}
}

\newcommand{\todo}[1]{\textbf{\color{red} #1}} 

\maketitle

\makeatletter
\let\ftype@table\ftype@figure
\makeatother

\begin{abstract}

We consider a classifier whose test set is exposed to various perturbations that are not present in the training set. These test samples still contain enough features to map them to the same class as their unperturbed counterpart. Current architectures exhibit rapid degradation of accuracy when trained on standard datasets but then used to classify perturbed samples of that data. To address this, we present a novel architecture named DeepConsensus that significantly improves generalization to these test-time perturbations. Our key insight is that deep neural networks should directly consider summaries of low and high level features when making classifications. Existing convolutional neural networks can be augmented with DeepConsensus, leading to improved resistance against large and small perturbations on MNIST, EMNIST, FashionMNIST, CIFAR10 and SVHN datasets.

\end{abstract}

\section{Introduction}

Classification algorithms are often trained with the goal of inferring samples as if they come from the training distribution. However, there is little guarantee that the training set forms an adequate support for the entire distribution. A common approach to addressing this is to augment the training set with artificial, anticipated transformations. However, there exist many practical and theoretical problems with this approach, including increased training time, model parameters and assumptions about the deployment conditions \cite{local-sicnn,ti-jungles,ti-pooling}. When faced with samples that are poorly represented in the training set, current state-of-the-art deep neural network architectures have built-in mechanisms that detect these as out-of-distribution (OOD) samples or suffer erroneous classifications with high confidence \cite{ood-confidence,ood-gan,ood-ensemble}. Although lower layers can successfully report the presence of invariant features in a perturbed image, their locations and orientations may not be well-modelled by higher layers. Only the last layer directly contributes to the final prediction in most current architectures, creating a single point of failure. We hypothesize that classifications based on a consensus of predictions made from both lower and higher layers will be more robust to perturbations without the need of augmenting training data. 

Convolutional neural networks (CNNs) and residual networks (ResNets) combine features locally using kernels of shared weights per layer and rely on increasingly abstract features with depth. ResNets use skip connections between blocks of convolutional layers to attain better performance with deeper architectures \cite{resnet}. In both architectures, each layer produces a convolution block consisting of a number of channels $C$. We interpret these blocks from a planar perspective such that each position on the plane contains a vector of length $C$. We refer to vectors from shallow layers as `lower level' features and those from deeper layers as `higher level' features. 

Our proposed architecture summarizes low and high level features across deep networks and uses consensus between these summaries to make classifications. The benefits of DeepConsensus include:
\begin{enumerate}
    \item Robustness against a variety of perturbations without training data augmentation (Figures \ref{allresults-incperturb}, \ref{deepfool-qualitativeresults}, Table \ref{deepfool-quantitativeresults}).
    \item The ability to learn sensitivity to properties that it is normally invariant to (Table \ref{mnist-40class-results}, examples in Figure \ref{mnist-40class-examples}). 
    \item Ease of attachment to a variety of existing architectures (Figure \ref{base-architectures}).
\end{enumerate}

\begin{figure*}[h]
    \centering
    \includegraphics[scale=0.5]{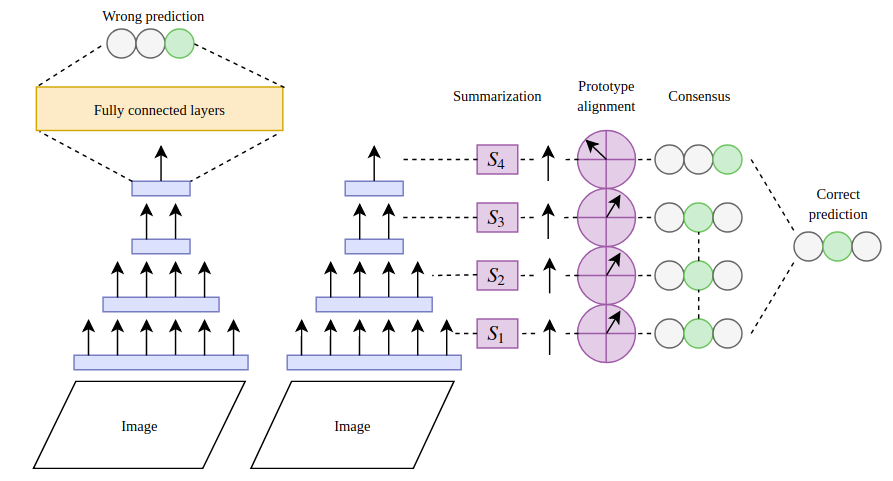}
    \caption{CNNs and ResNets \textit{(left)} rely on channel features \textit{(arrows)} from the highest layer. In contrast, DeepConsensus \textit{(right)} compares summaries of each layer to class prototypes and the final prediction arises from the consensus between multiple layers. The network is trained end-to-end with standard backpropagation. Purple denotes parameters involved only in DeepConsensus; orange denotes parameters involved only in the original network; blue boxes denote the layers that both architectures have in common; green and gray circles denote positive and negative predictions, respectively. We will directly observe the effect of replacing terminal fully connected layers with DeepConsensus in the experimental section.}
    \label{deepconsensus-diagram}
\end{figure*}

\section{Related work}

One way to achieve generalization is to model equivariant properties by representing changes in lower level features with similar changes in higher level features \cite{gconvs}. Another approach is to become invariant to these properties, which is necessary when data is scarce. There exist many architectures that possess equivariance or invariance towards specific, engineered properties. Rotation equivariant vector field networks apply filters orientated across a range of rotations \cite{roteq-vec}. Steerable filters have been successfully applied to CNNs to enable translational and rotational invariance simultaneously \cite{steerable-filters}. Spherical CNNs use kernels of shared weights on spherically projected signals and exhibit rotational equivariance \cite{spherical-cnn}. Scale-invariant CNNs use multiple columns \cite{sicnn} or kernels \cite{local-sicnn} of convolutional layers that specialize at different magnifications. Transformation invariant pooling uses Siamese CNNs that analyze two different transformations of the same object and selects the maximum outputs as the defining features of that class \cite{ti-pooling}. Group equivariant CNNs use special convolution filters that are capable of representing the equivariance of combinations of transformations. They perform well on the \textit{p4m} group transformations, achieving excellent scores on several datasets \cite{gconvs}. DeepConsensus differs from these models in two distinct ways: firstly, it does not require preemptively augmenting the training set and secondly, it is not engineered towards any particular type of perturbation or disturbance.

Another line of related work is OOD detection, where models are trained to produce low confidence scores on samples that are not from the training distribution. DeVries and Taylor proposed using incorrectly predicted training examples to learn a secondary confidence score classifier \cite{ood-confidence}. Vyas \textit{et al.} trained an ensemble of classifiers on different partitions of the training set to predict the other partition as OOD and used their agreement of OOD scores during evaluation \cite{ood-ensemble}. Lee \textit{et al.} used generative adversarial networks to learn edge examples of the training distribution to use for OOD classification \cite{ood-gan}. ODIN exploits the observation that small perturbations have a greater effect on temperature-scaled softmax predictions for in-distribution samples than OOD \cite{odin}. DeepConsensus does not aim to detect OOD samples explicity, but seeks to correctly classify samples with similar features to those seen in the training set. 

Prototypes are learnable representations in the form of one or more latent vector per class. Comparing features to prototypes instead of forming predictions directly from the features has shown promise in robust classification by several works \cite{prototype-fewshot, prototype-robust}. We build on the idea by using prototypes of feature summaries for every layer rather than only the deepest layer.

Zero- and few-shot learning are meta-learning concepts that aim to construct new categories for objects that do not exist in the training set, given meta-information or few examples. A classic zero-shot example is learning to classify zebras given a training set containing horses and semantic descriptions of zebras in relation to horses. DeepConsensus is different in that it seeks to recognize unfamiliar objects as having similarities to classes learned during training without the need of extra information.

\section{Architecture}\label{architecture_section}

\begin{figure}[h]
    \centering
    \includegraphics[scale=0.25]{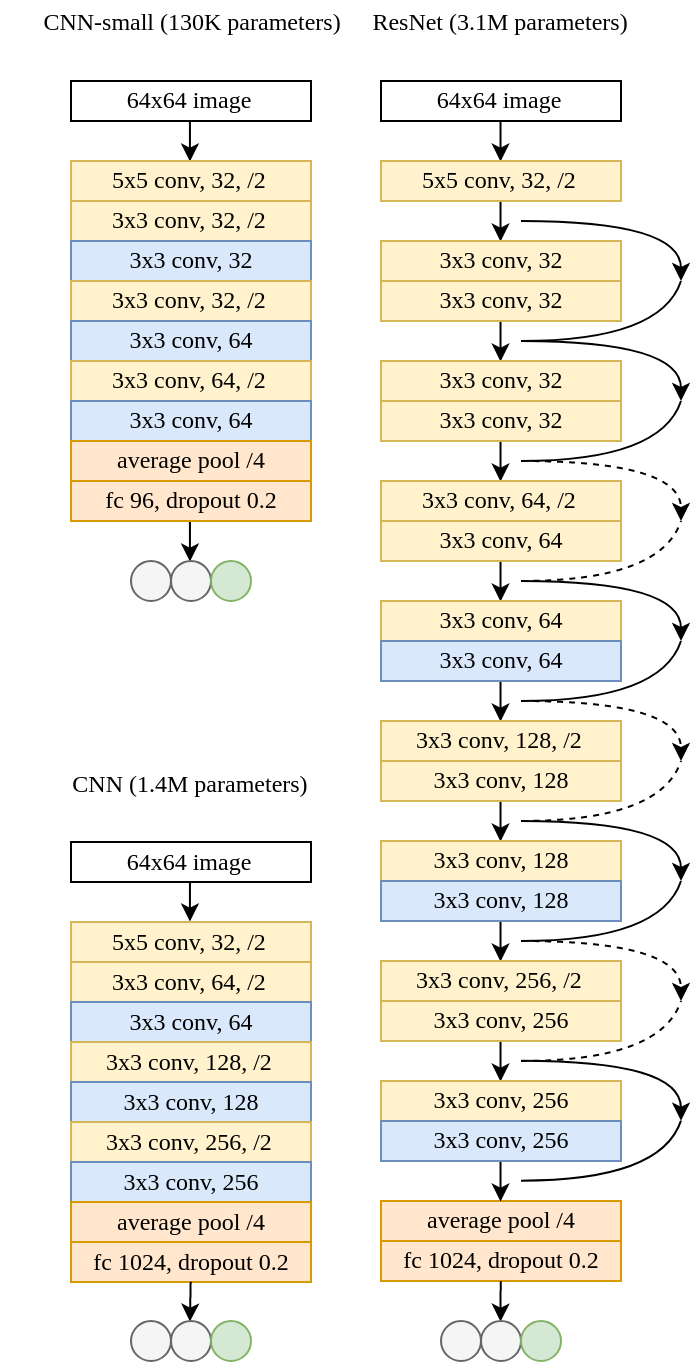}
    \caption{The CNN and ResNet architectures used for experiments. Boxes show kernel size, channel depth, and maxpooling size reduction, if present. Fully connected layers are denoted as \textit{fc}, followed by the number of hidden units. Blue boxes denote the convolution layer outputs sampled by DeepConsensus. Together with yellow boxes, they denote the layers required by both the original network and DeepConsensus. Orange boxes denote the additional layers required by the original networks, which are absent in DeepConsensus; gray and green circles denote example negative and positive class predictions; For ResNet, dotted arrows denote convolutions with size 1 kernels using a stride of 2 and all curved arrows represent addition, which are standard practices of the architecture \cite{resnet}. Batch norm follows every convolution and all activations are leaky ReLU with $\alpha=0.01$. After adding DeepConsensus and removing the fully connected layers, the parameter counts become 130 thousand for DeepConsensus-CNN-small, 1.2 million for DeepConsensus-CNN and 3.1 million for DeepConsensus-ResNet.}
    \label{base-architectures}
\end{figure}

The goal of DeepConsensus is to summarize outputs from each layer of a deep network such as CNN or ResNet, then compare the summaries to prototype summaries for each class, and finally find a consensus among these results for the prediction. Figure \ref{deepconsensus-diagram} gives a graphical overview of the architecture.

\subsection{Summarization}

In DeepConsensus, the summary operation $S_l$ of layer $l$ is defined as:
\begin{equation}
    S_l(x_l) = \sum_{i, j}{h_l\left(x_{l,i,j}; \theta_l\right)} = \tilde\sigma_l
\end{equation}
where $x_{l,i,j}$ represents the channel vector at row $i$ and column $j$ in the convolutional block of layer $l$ and $h_l:\mathbb{R}^d\to\mathbb{R}^{d'}$ is a nonlinear function with learnable parameters $\theta_l$. 

\subsection{Prototype alignment}

Summary vectors $\tilde\sigma_l$ are then compared to learned, layer-specific prototypes $\{\hat\sigma_{l,1}, \hat\sigma_{l,2}, ..., \hat\sigma_{l,c}, \hat\sigma_{l,c+1}\}$ using distance function $D_l:\mathbb{R}^{d'}\to\mathbb{R}^{c+1}$:
\begin{equation}
    D_l(\tilde\sigma_l) = [\delta(\tilde\sigma_l, \hat\sigma_{l,1}), \ldots, \delta(\tilde\sigma_l, \hat\sigma_{l,c}), \delta(\tilde\sigma_l, \hat\sigma_{l,c+1})]
\end{equation}
where $c$ denotes the number of classes and $\delta$ denotes some distance metric. The extra prototype allows layers to opt-out if the summary vector does not match any class prototype. We choose cosine similarity for the distance metric instead of Euclidean distance \cite{prototype-fewshot,prototype-robust}  because it is the most robust to the perturbations we tested (Figure \ref{ablation-result}).

\subsection{Consensus}

The consensus of predictions made by participating layers forms the final classification. For CNN or ResNet layer outputs $\{x_1,x_2...x_n\}$, DeepConsensus can be succinctly represented as the classification function $\phi$:
\begin{equation}
    \phi(x_1,x_2...x_n) = \sum_{l=1}^{n}w_l D_l(S_l(x_l))
\end{equation}
where $w_l$ weighs the contribution of layer $l$ to the final prediction.
Since conventional architectures use only the highest level features for classification, they can be expressed in the same form with weights:
$$w_l = 
    \begin{cases} 
      0 & l < n \\
      1 & l = n 
   \end{cases}
$$
and the distance metric is dot product.
In DeepConsensus, more than one layer makes a non-zero contribution to the final prediction and the distance metric is cosine similarity.

\subsection{Other architectural and training details}
The CNN and ResNet architectures used in the experiments are detailed in Figure \ref{base-architectures}. These networks are intended to demonstrate that DeepConsensus improves robustness of a variety of architectures as opposed to being tuned towards any particular one. All weights are initialized randomly from $\mathcal{N}(\mu=0,\sigma=0.02)$. Max-pooling is used instead of convolutions with stride 2 because of improved performance on perturbed test sets after training on unaltered training sets. These networks form the base network for DeepConsensus. We choose $h_l(\cdot)$ to be a single linear layer with a square weight matrix, followed by leaky ReLU with $\alpha=0.01$. Therefore, the prototypes of a particular layer have the same length as the number of channels of that layer. With this configuration, since DeepConsensus does not use terminal fully connected layers, the number of parameters of each network and its DeepConsensus version are practically equal.

The training regime is simple and standard, consisting of partitioning $20\%$ of the training set for validation, using cross entropy loss with Adam optimization \cite{adam} and reducing learning rate upon validation score plateau with factor 0.1 and patience 3. Since validation scores agree with unperturbed test scores and both values do not improve by more than 1\% of the total accuracy after 15 epochs of training, we choose 30 epochs to be the termination point at which actual test scores are taken.

\section{Experiments}

\begin{figure}[h]
    \centering
    \begin{tabular}{cc}
    $(a)$ & \subfloat{\includegraphics[valign=c,scale=0.5]{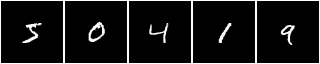}} \\ [-0.19in]
    $(b)$ & \subfloat{\includegraphics[valign=c,scale=0.5]{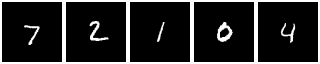}} \\ [-0.19in]
    $(c)$ & \subfloat{\includegraphics[valign=c,scale=0.5]{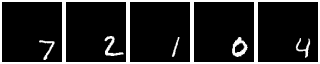}} \\ [-0.19in]
    $(d)$ & \subfloat{\includegraphics[valign=c,scale=0.5]{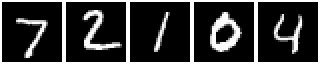}} \\ [-0.19in]
    $(e)$ & \subfloat{\includegraphics[valign=c,scale=0.5]{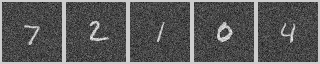}} \\ [-0.19in]
    $(f)$ & \subfloat{\includegraphics[valign=c,scale=0.5]{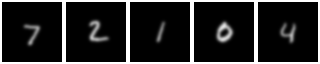}} \\
    \end{tabular}
    \caption{$(a)$ Training and validation examples, $(b)$ original test examples, and perturbations on the test set: $(c)$ translation of 20 pixels in $x$ and $y$ directions, $(d)$ magnification scale of 2, $(e)$ Gaussian noise addition with standard deviation of 30 and $(f)$ Gaussian blur with standard deviation of 1.5.}
    \label{mnist-traintestsamples}
\end{figure}

\begin{figure*}[h]
    \centering
    \includegraphics[scale=0.5]{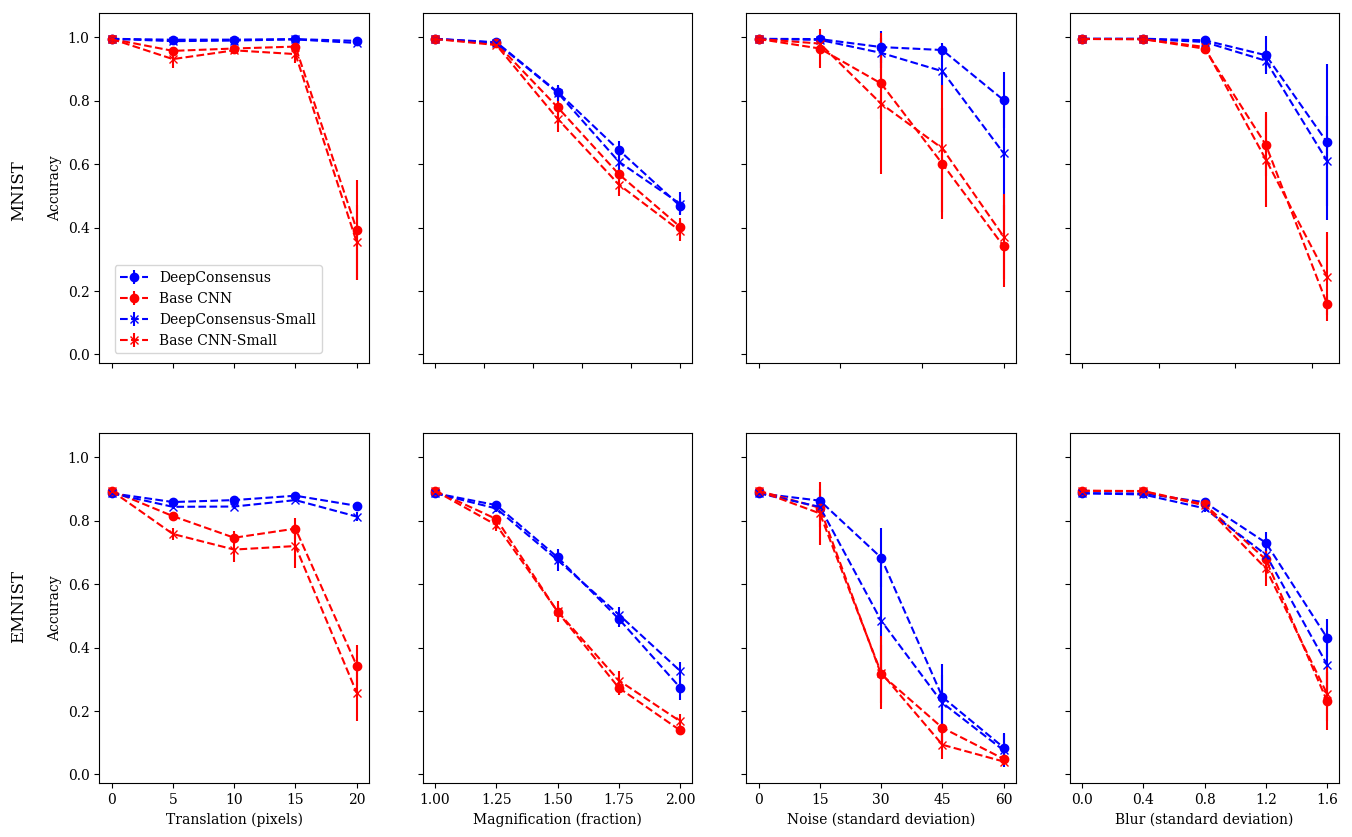}
    \caption{Robustness of CNN architectures improves with DeepConsensus across a range of translations, magnifications, Gaussian noise addition and Gaussian blurring. Data points represent means of 10 initializations and error bars represent 1 standard deviation. Hyperparameters are held constant across all conditions. Models are trained and validated on standard training sets, then tested on the corresponding perturbed test sets. Two models, Base CNN (1.4 million parameters) and Base CNN-Small (130k parameters) are considered along with their DeepConsensus variants. We note that DeepConsensus improves robustness regardless of network size. EMNIST with balanced 47 classes shows no degradation in performance with an increase in the number of distinctive classes. When the training set is perturbed similarly to the test set, all models achieve scores that are consistent with those observed in the absence of perturbations.}
    \label{allresults-incperturb}
\end{figure*}

\begin{figure}[h]
    \centering
    \includegraphics[scale=0.5]{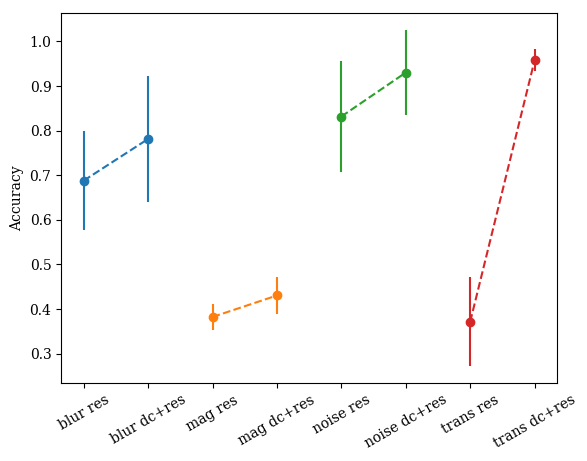}
    \begin{tabular}{lcc}
     \toprule
     Condition & T-statistic & P-value \\
     \midrule
     Translation & $40.34$ & $1.73\times10^{-42}$\\
     Magnification & $6.57$ & $3.37\times10^{-9}$ \\
     Noise & $4.43$ & $2.61\times10^{-5}$ \\
     Blur & $3.68$ & $3.85\times10^{-4}$ \\
     \bottomrule
    \end{tabular}
    \caption{\textit{(Top)} ResNet \textit{(res)} significantly improves in robustness with DeepConsensus augmentation \textit{(dc+res)} against robustness to translations (\textit{trans}, 20 pixels), magnification (\textit{mag}, 2x), Gaussian noise (\textit{noise}, 30 standard deviation) and Gaussian blur (\textit{blur}, 1.5 standard deviation). Data points represent means of 50 random initializations and error bars represent 1 standard deviation. \textit{(Bottom)} T-statistics and P-values are calculated using two-tailed independent T-tests with unequal variance.}
    \label{resnet-50x}
\end{figure}

We train and validate the base CNN, ResNet, and their corresponding DeepConsensus augmentations on the standard training samples of MNIST \cite{mnist}, EMNIST with a balanced split of 47 classes \cite{emnist}, FashionMNIST \cite{fashion} and CIFAR10 \cite{cifar}. To accommodate the spatial perturbations on the test set, both training and test samples are placed in the center of a $64\times64$ black background. The training set is not perturbed, while the test samples are subject to increasing levels of translation, magnification (using nearest-neighbor interpolation), addition of Gaussian noise, and blurring. Examples of the MNIST training and testing sets are shown in Figure \ref{mnist-traintestsamples} and results are shown in Figure \ref{allresults-incperturb} and Supplementary Figure \ref{results-fashioncifar}. Figure \ref{resnet-50x} demonstrates similar improvements with the ResNet architecture. DeepConsensus does equally well on the unperturbed test set as its base network, but also develops invariance to translation and moderate resistance against the other conditions even without training set augmentation. Table \ref{variance-after-fixedinit} shows that the high variance in scores on heavily perturbed test sets is mainly due to parameter initialization.

\begin{figure}[h]
    \centering
    \includegraphics[scale=0.5]{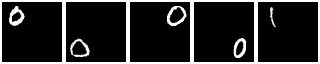}
    \caption{Examples of the classes $0$ to $4$ for MNIST quadrants. This test measures spatial sensitivity.}
    \label{mnist-40class-examples}
\end{figure}

\begin{table}[h]
    \centering
    \caption{Test accuracy on MNIST quadrants, where the spatial position of the digit also determines its class. The hyperparameters of these models are held constant from the perturbation study and the results are repeated 3 times with random initializations. Despite being invariant to translation, DeepConsensus can still do well on spatial tasks.}
    \begin{tabular}{lc}
    \toprule
    Model & Test score \\
    \midrule
    Base CNN & $0.9951 \pm 0.0004$ \\
    DeepConsensus & $0.9957 \pm 0.0003$\\
    \bottomrule
    \end{tabular}
    \label{mnist-40class-results}
\end{table}

\begin{table}[h]
    \centering
    \caption{The largest contributor to variance in test scores is parameter initialization. Here we show the effect of using different initializations versus reusing the first initialization of 10 repeated trials on test score standard deviation. The test condition is Gaussian blur with standard deviation of 1.6, which shows the highest variance (see Figure \ref{allresults-incperturb}).}
    \begin{tabular}{lccc}
        \toprule
        Model & Random & Fixed & Decrease \\
        \midrule
        Base CNN & $0.14$ & $0.043$ & 69.3\%\\
        DeepConsensus & $0.14$ & $0.015$ & 89.3\%\\
        \bottomrule
    \end{tabular}
    \label{variance-after-fixedinit}
\end{table}

\begin{figure*}[h]
    \centering
    \includegraphics[scale=0.5]{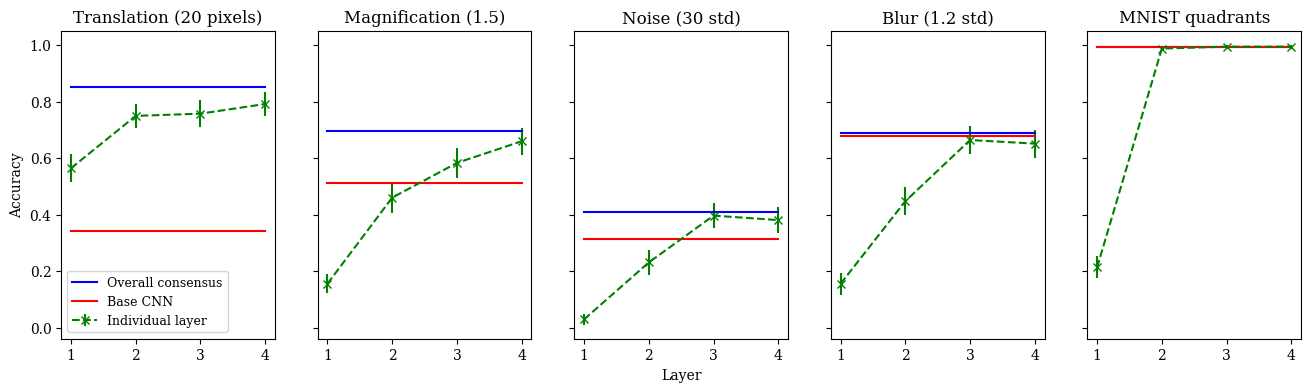}
    \caption{Test accuracy of each individual layer is different depending on the perturbation type (Gaussian noise, blurring, translation and magnification) on EMNIST and on the synthesized 40-class MNIST quadrants. Data points represent means of 10 random initializations and error bars represent 1 standard deviation. Whereas the training data of the EMNIST experiments do not contain the perturbations made on the test sets, MNIST quadrants had independent and identically distributed training and test sets. Note that overall consensus is of greater accuracy than its constituents and consensus improved individual layer performance over the base network in some conditions. We use DeepConsensus-ResNet modified with an additional consensus layer at the 5th convolution to obtain accuracies of 4 layers.}
    \label{layerwise-accuracy}
\end{figure*}

Since DeepConsensus appears to be immune to large translation perturbations, we questioned if it is capable of classification tasks that depend on spatial positioning. We synthesized a 40 class MNIST dataset where we placed the original $28\times28$ image randomly in one of the four quadrants of a $64\times64$ black image and maintained the same training set (60k) and test set (10k) size. Each digit is mapped to 4 different classes corresponding to their quadrant location (see Figure \ref{mnist-40class-examples}). Table \ref{mnist-40class-results} shows that DeepConsensus achieves the same test score as its base network, demonstrating sensitivity to spatial positioning despite being normally invariant to translation. Figure \ref{layerwise-accuracy} shows the average contribution of each layer on the previous experiments and this dataset.

\begin{figure}[h]
    \centering
    \includegraphics[scale=0.45]{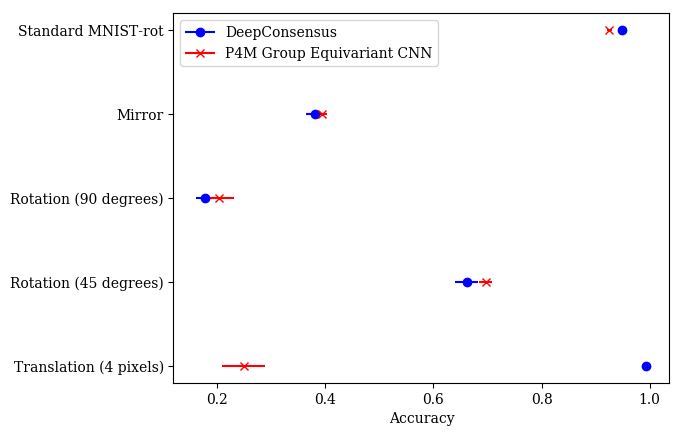}
    \caption{$p4m$-CNNs are capable of equivariance for translations, rotations, and mirror reflections. They produce state-of-art results on the standard MNIST-rot dataset. Unlike DeepConsensus, they require a training set with similar perturbations as the test set. We train the models for 30 epochs on $80\%$ of the training set and validate on the reserved $20\%$ portion. We show final test scores on the standard MNIST-rot dataset, as well as MNIST variants where the training set is not perturbed. Data points represent means for 3 random initializations and error bars represent 1 standard deviation. The mirror variant is a horizontal reflection across the central $y$-axis. We use the state-of-art $p4m$-CNN architecture on MNIST-rot. }
    \label{p4m-results}
\end{figure}

Conventional CNNs and ResNets are invariant to translation locally and achieve equivariance to global translation only if the training set is perturbed in a similar way to the test set, as demonstrated by the translation condition in Figure \ref{allresults-incperturb}. As Figure \ref{layerwise-accuracy} shows, DeepConsensus uses consensus to exploit the soft prediction scores from each layer, such that overall accuracy is greater than that of any one particular layer. On the other hand, conventional networks use higher-level features exclusively, which are spatially-weighted combinations of lower level features and are not well-modelled for perturbed inputs, causing low accuracy on such tasks. We compare DeepConsensus to the state-of-art architecture \textit{p4m}-CNN \cite{gconvs} on the standard MNIST-rot dataset, where both the 12K training and 50K test samples are rotated randomly about the center axis \cite{mnist-rot}. Furthermore, we compare both models on variations of MNIST where the training set is unaltered but the test set is perturbed. We demonstrate that DeepConsensus performs similarly to $p4m$-CNNs at perturbation tasks the $p4m$-CNN model is specialized for, and significantly outperforms it in translation (Figure \ref{p4m-results}).

\begin{figure*}[h]
    \centering
    \includegraphics[scale=0.5]{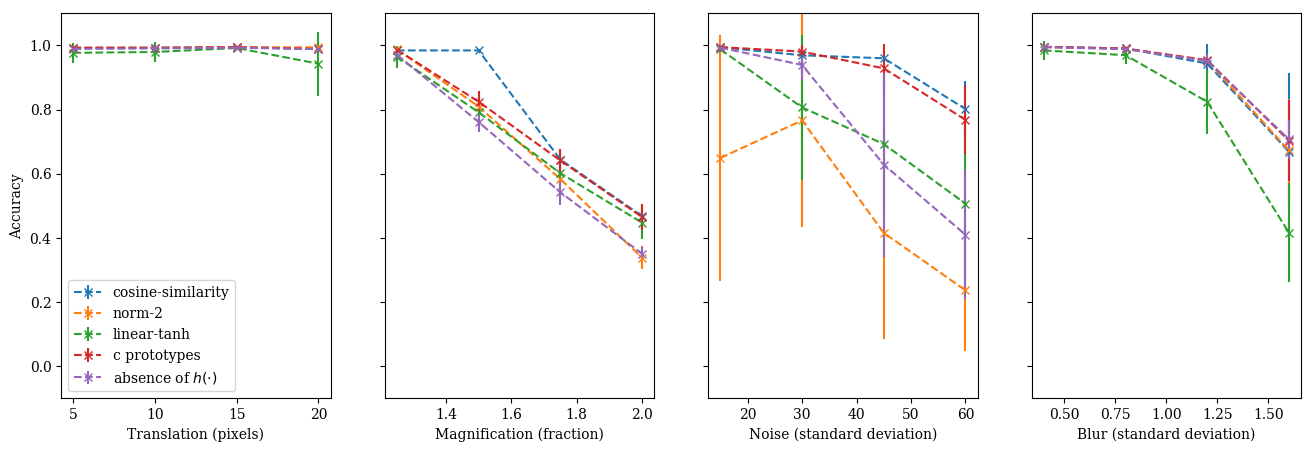}
    \caption{Ablation studies of DeepConsensus on MNIST with perturbations on the test set only. Data points represent means of 10 random initializations and error bars represent 1 standard deviation. Results illustrate the effect of using cosine similarity over Euclidean distance or fully connected layers as the distance function, using $c$ instead of $c+1$ prototypes, and using a nonlinear transformation $h(\cdot)$ before summarizing the convolution layer.}
    \label{ablation-result}
\end{figure*}

We perform ablation studies on DeepConsensus (Figure \ref{ablation-result}). Best results are achieved when using a nonlinear transformation on features before summarization, and using cosine similarity as the metric for comparison with prototypes.

\begin{figure}[h]
    \centering
    \begin{tabular}{cccc} \
    \subfloat{\includegraphics[scale=0.5]{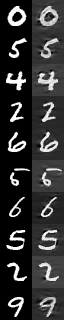}} &
    \subfloat{\includegraphics[scale=0.5]{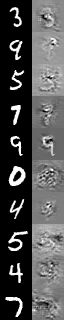}} &
    \subfloat{\includegraphics[scale=0.5]{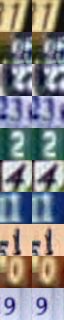}} &
    \subfloat{\includegraphics[scale=0.5]{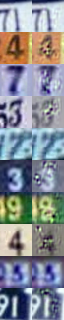}} \\
    $(a)$ & $(b)$ & $(c)$ & $(d)$ \\
    \end{tabular}
    \caption{Greater perturbation of the input is necessary to fool models enhanced with DeepConsensus. For each paired column, the left shows original samples while the right shows the corresponding perturbed sample found by DeepFool that forces the model to choose a different class. Random adversarial examples for ResNet ($a$, $c$) and DeepConsensus-ResNet ($b$, $d$) are collected after training on standard MNIST ($a$, $b$) and SVHN ($c$, $d$) for 30 epochs.}
    \label{deepfool-qualitativeresults}
\end{figure}

Having observed that DeepConsensus is robust against large perturbations, we subject DeepConsensus-ResNet to adversarial examples to observe its robustness against small perturbations. After training on MNIST and SVHN \cite{svhn} for 30 epochs, both the base ResNet and the DeepConsensus-ResNet version are analyzed with DeepFool, which attempts to find the smallest perturbation on the input to force the model to change classification output \cite{deepfool}. The adversarial robustness metric or perturbation density is
\begin{equation}
    \hat{\rho}(f) = \frac{1}{|\mathcal{D}|}\sum_{x\in\mathcal{D}}\frac{||\hat{r}(x)||_2}{||x||_2}
\end{equation}
where $f$ is the classifier, $\mathcal{D}$ is the test set, and $\hat r(x)$ is the minimal perturbation found by DeepFool \cite{deepfool}.
Perturbation densities are significantly higher with DeepConsensus (Table \ref{deepfool-quantitativeresults}), showing that DeepConsensus is robust against the small, targeted perturbations of DeepFool. Figure \ref{deepfool-qualitativeresults} compares sample adversarial examples for DeepConsensus-ResNet versus the base ResNet on MNIST and SVHN. 

\begin{table}[h]
    \centering
    \caption{Perturbation densities from DeepFool increase with the addition of DeepConsensus to the base ResNet. A larger adversarial perturbation signifies a more robust model against small perturbations. 1000 samples are randomly chosen from each test set for adversarial analysis.}
    \begin{tabular}{lccc}
    \toprule
    Dataset & ResNet & DeepConsensus & Increase \\
    \midrule
    MNIST & $0.180 \pm 0.065$ & $2.9 \pm 1.7$ & 1500\%\\
    SVHN & $0.015 \pm 0.012$ & $ 0.12 \pm 0.12$ & 700\%\\
    \bottomrule
    \end{tabular}
    \label{deepfool-quantitativeresults}
\end{table}

The code to produce these results is found \href{https://github.com/ychnlgy/DeepConsensus-experimental-FROZEN}{here}. A user-friendly version of the code is found \href{https://github.com/ychnlgy/DeepConsensus}{here}.

\section{Discussion}

In the experiments, we show that DeepConsensus improves the robustness of different architectures against multiple types of perturbation. Another major strength of DeepConsensus is the ability to become sensitive to spatial properties despite normally being invariant to them (e.g., Table \ref{mnist-40class-results}). This is because it uses high-level features when they agree with lower level features, or take a consensus of partially correct predictions when their first choice predictions (FCP) do not align. As Figure \ref{layerwise-accuracy} shows, when trained on MNIST quadrants, the features of the last three layers are well aligned and correct. Therefore, the FCP of the sum of their predictions is equal to that of any one of their predictions. In contrast, for EMNIST with perturbations on the test set, the higher accuracy exhibited by consensus is similar in mechanism to an ensemble of classifiers -- although the FCP for any one particular layer may be incorrect, the FCP of their sum is more likely to be correct. It is precisely the agreement of soft prediction scores that allows DeepConsensus to become sensitive, or invariant, to spatial positioning.

We designed DeepConsensus with the intention of achieving magnification invariance. Computing the cosine similarity between summaries of features and prototypes is equivalent to checking if the two vectors have some approximate scalar relation. Effectively, this operation finds the class whose prototype contains a similar ratio of features to the input. We originally believed that magnification maintains the ratio of features, but two observations falsify this belief: increasing the image size decreases the border width in a non-proportional way, and features are weighted differently from one another. We tried applying various bounded activation functions to mitigate these effects, but were not successful. The network assigns different weightings to different features regardless of activation expressiveness, preventing ratio-based prototype comparisons from becoming fully invariant to magnification. 

The ablation studies (Figure \ref{ablation-result}) suggest cosine similarity to be the most robust for comparing layer summaries to class prototypes. Using Euclidean distance for the comparison metric does not perform well for magnification nor Gaussian noise addition, which have different ratios of features compared to the other perturbation types. This is consistent with the optimal solution of Euclidean-based prototype matching, which is to find a class prototype that matches the sample vector exactly. On the other hand, cosine similarity matches only the ratio of features, so it is a less rigid matching function that is more suitable for robust classification. Using a linear layer to classify the summary vectors leads to poor performance across all four perturbations, suggesting that fully connected layers of classifiers should be replaced with prototype-based comparisons to improve general robustness. 

The perturbation densities of adversarial examples found by DeepFool are higher for models augmented with DeepConsensus (Table \ref{deepfool-quantitativeresults}). Although the intention of this experiment is to demonstrate that DeepConsensus is resistant to small perturbations and we do not claim DeepConsensus to be robust against adversarial attacks, it illuminates a potential direction in adversarial defence. Instead of finding ways of preventing adversarial examples from being found, which may not be possible, we should aim to develop models that can directly consider higher and lower level features. Simply depending on higher level features creates a single point of failure at the deepest layer, whereas to fool all layers of an entire network may result in large and more obvious adversarial perturbations.

\subsection{Future work}

Closely related to the idea of few-shot learning, a practical application of DeepConsensus is to take a small, supervised batch of the test distribution to determine the accuracy of each layer. These accuracy can then be multiplied as the weighted contribution of each layer for the model to dynamically adapt to different deployment conditions without the need of retraining.

Follow up work includes showing changes in performance on more drastic perturbations, such as broad non-linear rasterizations and obstruction of distinguishing characteristics. Additionally, further research is needed on how DeepConsensus can:
\begin{enumerate}
    \item be adapted to work with large, pretrained models, like the Inception versions or deep ResNet networks, where consensus is built along their layers,
    \item augment existing invariant models, such as improving translational invariance of $p4m$-CNNs, and
    \item attain generalization when training data is scarce.
\end{enumerate}
We also wish to find a better summarization approach than simply summing features. We originally tried a more sophisticated approach where the contributions of layers depended on the agreement between their predictions and the predictions of their neighboring layers. This was not successful because deeper layers tended to have strong agreement regardless of their accuracy, producing similar behavior to the base network on perturbed samples.

\section{Conclusion}

We expose weaknesses of various models on classification of perturbed samples, when trained on standard training sets. We propose augmenting existing models with the DeepConsensus architecture to improve their resistance to a variety of perturbations. DeepConsensus is not expensive in terms of parameters and does not require any preemptive training set augmentation. We also show that it is amenable to classification tasks that are sensitive to spatial positioning, and that each of its components are necessary and effective at improving general robustness. In addition, we demonstrate the resistance of DeepConsensus against the small, targeted perturbations of DeepFool.  

\clearpage

\section{Acknowledgements}

Special thanks to Kamyar Ghasemipour, who provided important suggestions and helpful discussions.

{\small
\bibliographystyle{ieee}
\bibliography{egbib}
}

\clearpage
\onecolumn

\setcounter{figure}{0}

\section{Supplementary material}

\begin{figure*}[h]
    \centering
    \includegraphics[scale=0.45]{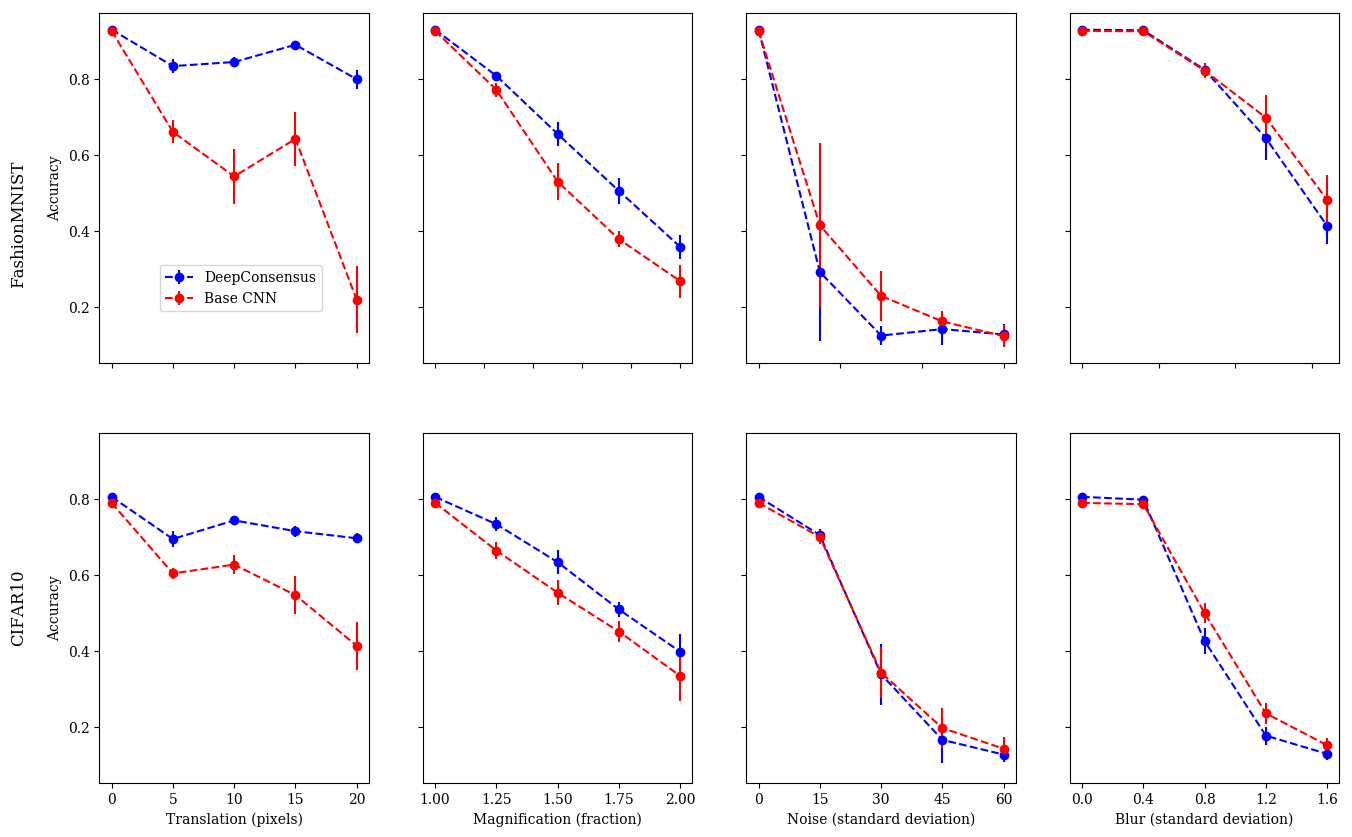}
    \caption{For FashionMNIST and CIFAR10, the difference between test scores on translation and magnification perturbation is more pronounced than those seen in MNIST and EMNIST, but we observe decreased overall performance with Gaussian noise and blurring. Data points represent means of 10 random initializations and error bars represent 1 standard deviation. Hyperparameters are held constant from the MNIST and EMNIST experiments. }
    \label{results-fashioncifar}
\end{figure*}

\end{document}